\documentclass[runningheads]{llncs}
\usepackage{graphicx}

\usepackage{tikz}
\usepackage{comment}
\usepackage{amsmath,amssymb} 
\usepackage{color}
\usepackage{soul}
\usepackage[colorlinks,
            linkcolor=red,      
            anchorcolor=red,  
            citecolor=blue,       
            ]{hyperref}
\usepackage[width=112mm,left=17mm,paperwidth=146mm,height=177mm,top=20mm,paperheight=217mm]{geometry}

\begin{document}
\pagestyle{headings}
\mainmatter
\def\ECCVSubNumber{96}  

\title{Deep Relighting Networks for Image Light Source Manipulation} 


\titlerunning{Deep Relight. Networks for Image Light Source Manipulation}
%
\author{Li-Wen Wang\inst{1}\and
Wan-Chi Siu\inst{1} \and
Zhi-Song Liu\inst{2} \and\\
Chu-Tak Li\inst{1} \and
Daniel P.K. Lun\inst{1}
}
\authorrunning{L.W. Wang, W.C. Siu, et al.}
%
\vskip -1\baselineskip
\institute{The Hong Kong Polytechnic University, Hong Kong \and
LIX, Ecole Polytechnique, CNRS, IP Paris, France
}
\maketitle
\vskip -0.6cm
\vskip -2\baselineskip 
\begin{abstract}
Manipulating the light source of given images is an interesting task and useful in various applications, including photography and cinematography. Existing methods usually require additional information like the geometric structure of the scene, which may not be available for most images. In this paper, we formulate the single image relighting task and propose a novel Deep Relighting Network (DRN) with three parts: 1) scene reconversion, which aims to reveal the primary scene structure through a deep auto-encoder network, 2) shadow prior estimation, to predict light effect from the new light direction through adversarial learning, and 3) re-renderer, to combine the primary structure with the reconstructed shadow view to form the required estimation under the target light source. Experiments show that the proposed method outperforms other possible methods, both qualitatively and quantitatively. Specifically, the proposed DRN has achieved the best PSNR in the “AIM2020 - Any to one relighting challenge" of the 2020 ECCV conference.
 
\keywords{Image relighting, back-projection theory, deep learning}

\end{abstract}
\vskip -1cm
\vskip -2\baselineskip 
\section{Introduction}
\vskip -0.5\baselineskip 
Image is a popular information carrier in this information era, which is intuitive and easy to understand. The rapid development of display devices stimulates people's demand for high-quality pictures. The visual appearance of the images is highly related to the illumination, which is vital in various applications, like photography and cinematography. Inappropriate illumination usually causes various visual degradation problems, like undesired shadows and distorted colours. However, the light source (like sunlight) is difficult to control, or sometimes unchangeable (for captured images), which increases the difficulty of producing satisfying images.  The ways to produce the effect of light source on captured images becomes a hi-tech topic which has attracted considerable attention, because it offers opportunities to retouch the illuminations of the captured images.  

Some approaches have been proposed that aim to mitigate the degradation caused by improper illuminations. For example, histogram equalization (HE) \cite{HE_1} rearranges the intensity to obey uniform distribution, which increases the discernment of the low-contrast regions. It balances the illumination of the whole image that manipulates the global light condition.  Methods \cite{HDR97,HDR_ECCV} in the high-dynamic-range (HDR) field improve the image quality by increasing the dynamic range of the low-contrast regions. The HDR methods can be regarded as a refinement of local contrast but lacks adjustment of the global light. Retinex-based methods \cite{Retinex2004,RetinexNet} separate the images as the combination of illumination and reflectance, where the reflectance stores the inherent content of the scene that is unchangeable in different illumination conditions. By refining the illumination, it can improve the visual quality of the images.  Low-light image enhancement methods \cite{EnlightenGAN,DLN2020} amend the visibility of the dark environment that enlighten the whole image. Shadow removal \cite{shadow_2018,shadow_2019} is a popular topic in the field of image processing that aims to eliminate the shadow effects caused by the light sources, but cannot simulate the shadows for target light sources. Adjusting the light source provides a flexible and natural way for illumination-based image enhancement. Although considerable research has been devoted to refine the illumination, less effect is being made to study from the view of manipulating the light sources. In other words, changing the illumination by controlling the light source is still in its fancy stage. Literature in relighting field mainly focuses on specific applications, like portrait relighting \cite{Nestmeyer2020LearningPF,Sun2019SingleIP,portrait_relight_2019}. These methods require prior information (like face landmarks, geometric priors) that cannot be implemented in general scenes.

Convolutional Neural Network (CNN) recently has attracted notable attention due to its powerful learning capacity. It can digest extensive training data and extract discriminative representations with the support of powerful computational resource. CNN has shown significant advantages in various tasks, like image classification \cite{imagenet,VGG}, semantic segmentation \cite{seg2015,segYu}, super-resolution\cite{SR19,SR20}, place recognition \cite{ToDayGAN,CNNlocalization}, etc. CNNs with the deep structure are difficult to train because parameters of the shallow layers are often under gradient vanishing and exploding risks.  Residual learning \cite{Resnet2016} mitigates the optimizing difficulty by adding a shortcut connection among each processing block. With the assistance of the normalization layers, the gradient can flow from the deep to shallow layers steadily, which dramatically increases the training efficiency of the deep network. The deeper structure usually means more trainable parameters that bring in more powerful learning capacities, which makes it possible to handle more challenging tasks, like single image relighting.

\begin{figure}[t]
    \centering
    \includegraphics[width=\textwidth]{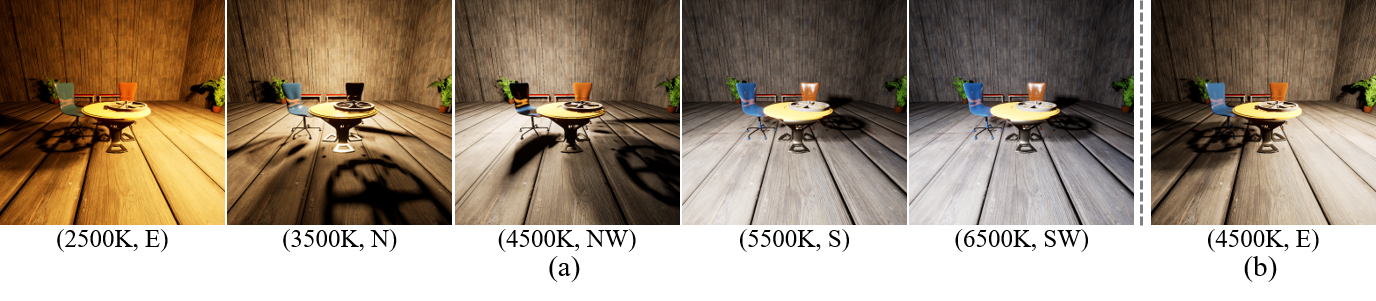}
   \vskip -2\baselineskip 
    \caption{An example of the ``any to one" relighting task. The 2500K, 3500K, ... are the color temperatures, and the E, N, ... are the light directions. Images in (a) are the inputs with any light settings, and (b) is target out with a specific light setting.}
    \label{fig:any2one}
   \vskip -1.5\baselineskip 
\end{figure}

The image relighting method in this paper focuses on manipulating the position and color temperature of the light source based on our powerful deep CNN architecture. It not only can adjust the dominant hue, but can also recast the shadows of the given images. As shown in Fig. \ref{fig:any2one}, we focus on a specific ``any to one" relighting task \cite{elhelou2020aim}, for which the input is under arbitrary light sources (any direction or color temperature, see Fig. \ref{fig:any2one}(a)), and the objective is to estimate the image under this specific light source (direction: E, color temperature: 4500K, see Fig. \ref{fig:any2one}(b)). The proposed method can be generalized for other light-related tasks. Let us highlight the novelty of our proposed approach.

\vskip -0.5\baselineskip 
\begin{itemize}
\item Instead of directly mapping the input image to the target light condition, we formulate the relighting task in three parts: scene reconversion, light effects estimation and re-rendering process. 
\item To preserve more information of the down- and up-sampling processes, we insert the back-projection theory to the auto-encoder structure, which benefits the scene reconversion and light-effect estimation.
\item The light effect is difficult to measure, which increases the training difficult. We use the adversarial learning strategy that is implemented by a new shadow-region discriminator, which gives guidance to the training process. 
\end{itemize}

\vskip -0.5cm
\vskip -1.5\baselineskip 
\section{Related Works}
\vskip -0.5\baselineskip 
\subsubsection{Back-Projection (BP) theory.}
BP theory is popular in the field of single-image super-resolution \cite{haris2018dbpn,Liu2019hbpn,Liu2019abpn}. Instead of directly learning the mapping from the input to the target, the BP-based methods iteratively digest the residuals and refine the estimations. It gives more focus on the weakness (i.e., the residuals) that appears at the learning process, which significantly improves the efficiency of the deep CNN architectures.

Recent work on low-light image enhancement \cite{DLN2020}  extends the BP theory to the light-domain-transfer tasks. It assumes the low-light (LL) and normal-light (NL) images locate at the LL and NL domains separately. Firstly, a lightening operator predicts the NL estimation from the LL input. Then, a darkening operator maps the NL estimation back to the LL domain (LL estimation). In the LL domain, the difference (LL residual) between LL input and LL estimation can be found that indicates the weakness of the two transferring operators (lightening and darkening). Afterwards, the LL residual is mapped back to the NL domain (NL residual) through another lightening operator. The NL residual then refines the NL estimation for a better output. Mathematically, the enlightening process can be written as:
\begin{align}
  \hat{\mathbf{N}} = \lambda_{2}L_{1}(\mathbf{L}) + L_{2}(D(L_{1}(\mathbf{L})) - \lambda_{1}\mathbf{L})
\end{align}
where $\mathbf{L}$ and $\hat{\mathbf{N}}$ $\in\mathbb{R}^{\mathrm{H} \times \mathrm{W} \times \mathrm{3}}$ denote the LL input image and NL estimation separately. The terms $H$, $W$ and $3$ represent the height, width and RGB channels respectively. The symbols $L_1$ and $L_2$ are two lightening operators to enlighten the LL image and LL residual individually. The symbol $D$ is the darkening operator that maps the NL estimation to the LL domain. Two weighting coefficients $\lambda_1$ and $ \lambda_2 \in \mathbb{R}$ are used to balance the residual calculation and final refinement.

\vskip -0.5cm
\vskip -1\baselineskip 
\subsubsection{Adversarial Learning.}
Transferring an image to a corresponding output image is often formed as a pixel-wised regressing task of which the loss function (like L1- or L2-norm loss) indicates the average error for all pixels. This type of loss functions neglects the cross-correlation among the pixels, which easily distorts the perceptual structure and causes blur outputs. A large number of research works have done on quantitative measures of the perceptual similarity among images, like  Structure SIMilarity (SSIM) \cite{SSIM}, Learned Perceptual Image Patch Similarity (LPIPS) \cite{LPIPS}, Gram matrix \cite{grammatri}, etc. However, the perceptual evaluation basically varies from different visual tasks and is difficult to formulate.

The Generative Adversarial Networks (GANs) \cite{GAN,pix2pix,cGAN} provide a novel solution that embeds the perceptual measurement into the process of adversarial learning. Each GAN consists of a generator and a discriminator. The discriminator aims to find latent perceptual structure inside the target images, which then guides the training of the generator. Subsequently, the generator provides sub-optimal estimations that will work as negative samples for the training process of the discriminator. With the grouped negative and positive (target images) samples, the discriminator conducts a binary classification task, which measures the latent perceptual difference between the two types of samples. The overall training process is shown as:
\begin{align}
  \min_{G}\max_{D}V(D,G)=\mathbb{E}_{\mathbf{Y}}[logD(\mathbf{Y})]+\mathbb{E}_{\mathbf{X}}[log(1-D(G(\mathbf{X})))]
\end{align}
where $D$ and $G$ denote the discriminator and generator separately. The terms $\mathbf{X}$ and $\mathbf{Y}$ represent the input and target images respectively. In the training process, the generator and discriminator play a two-player minimax game. The discriminator learns to distinguish the estimated images $G(\mathbf{X})$ from the target ones $\mathbf{Y}$. The generator aims to minimize the difference between the estimated $G(\mathbf{X})$ and target images $\mathbf{Y}$. The training process follows the adversarial learning strategies, where the latent distribution inside the target images is increasingly learned and used. Finally, the training will reach a dynamic balance, where the estimations produced by the generator have similar latent perceptual structure as the real target images.

\vskip -0.1cm
\vskip -0.5\baselineskip 
\section{The Proposed Approach}
\vskip -0.5\baselineskip 

\begin{figure}[t]
    \centering
    \includegraphics[width=0.9\textwidth]{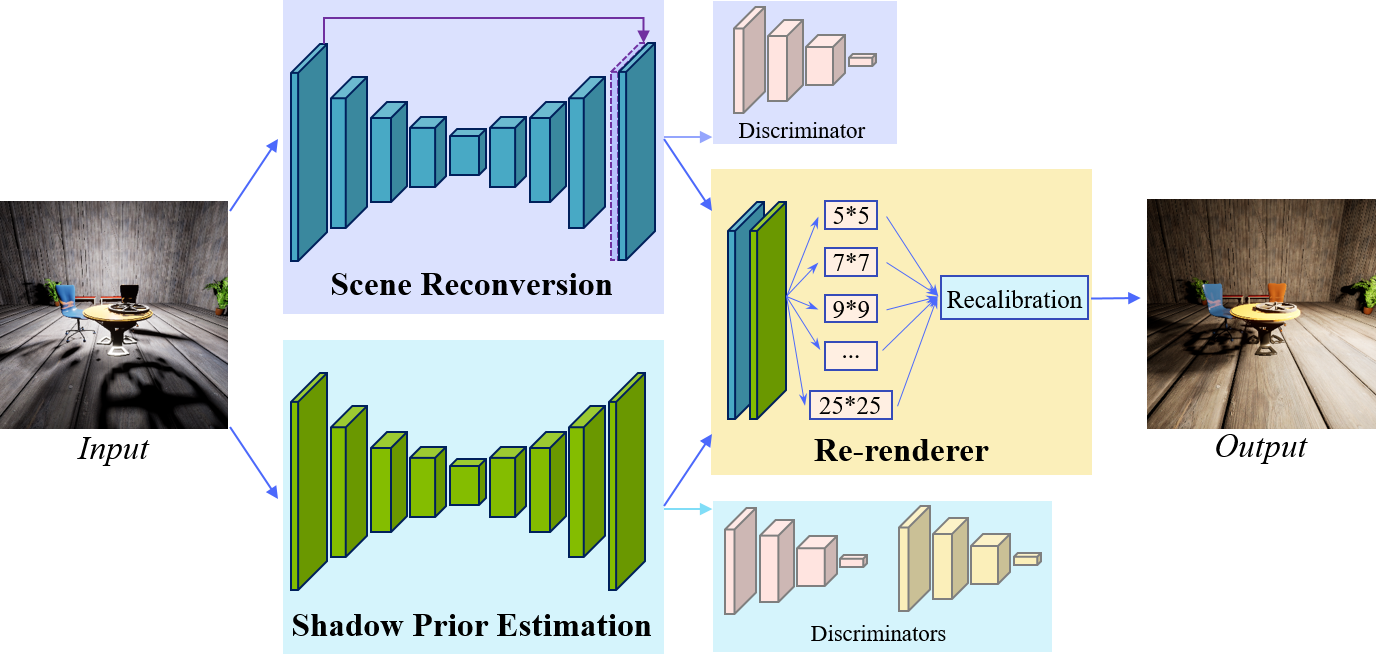}
   \vskip -1\baselineskip 
    \caption{Architecture of the proposed method}
    \label{fig:arch}
   \vskip -1.5\baselineskip 
\end{figure}

As shown in Fig. \ref{fig:arch}, the proposed Deep Relighting Network (DRN) consists of three parts: scene reconversion, shadow prior estimation, and re-renderer. Firstly, the input image is handled in the scene reconversion network (see Section \ref{Scene Reconversion}) to remove the effects of  illumination, which extracts inherent structures from the input image. At the same time, another branch (shadow prior estimation, see Section \ref{shadowprior}) focuses on the change of the lighting effect, which recasts shadows according to the target light source. Next, the re-renderer part (see Section \ref{rerender}) perceives the lighting effect and re-paints the image with the support of the structure information. Both the scene reconversion and shadow prior estimation networks have a similar deep auto-encoder structure that is an enhanced variation of the ``Pix2Pix" network \cite{pix2pix}. The details of the three components are presented below. 
\vskip -0.2\baselineskip 
\vskip -0.5\baselineskip 
\subsection{Assumption of Relighting}
\label{assumption}
\vskip -0.5\baselineskip 
Any-to-one Single image relighting is a challenging low-level vision task that aims to re-paint the input image $\mathbf{X}\in\mathbb{R}^{\mathrm{H} \times \mathrm{W} \times \mathrm{3}}$ (under any light source $\mathrm{\Phi}$) with the target light source $\mathrm{\Psi}$. Inspired by the Retinex theory \cite{Retinex2004,RetinexNet}, we assume that images can be decomposed into two components, where structure $\mathbf{S}$ is the inherent scene information of the image that is unchangeable under different light conditions. Let us define a lighting operation $L_{\mathrm{\Phi}}(\cdot)$, which provides global illumination and causes the shadow effects for the scene $\mathbf{S}$ under the light source $\mathrm{\Phi}$. The input image can be written as:
\vskip -0.4cm
\vskip -0.5\baselineskip 
\begin{align}
  \mathbf{X} = L_{\mathrm{\Phi}}(\mathbf{S})
\end{align}
\vskip -0.5\baselineskip 

To re-paint the image $\mathbf{X}$ with another light source $\mathrm{\Psi}$, it firstly needs to remove the lighting effect $L^{-1}_{\mathrm{\Phi}}(\cdot)$, i.e., reconverting the structure information $\mathbf{S}$ from the input image $\mathbf{X}$. Then, with the target light operation $L_{\mathrm{\Psi}}(\cdot)$, the image $\mathbf{Y}$  with the target light source can be obtained through:
\vskip -0.4cm
\vskip -0.5\baselineskip 
\begin{align}
  \mathbf{Y} = L_{\mathrm{\Psi}}(  L^{-1}_{\mathrm{\Phi}}(\mathbf{X} ) )
\end{align}
\vskip -0.5\baselineskip 

The key part of the reconversion process $L^{-1}_{\mathrm{\Phi}}(\cdot)$ is to eliminate the shadows, while the lighting operation $L_{\mathrm{\Psi}}(\cdot)$ is to paint new shadows for the target light source. However, the geometric information is unavailable in the single image relighting task, which dramatically increases the difficulty of constructing the lighting operation $L_{\mathrm{\Psi}}(\cdot)$. Hence, instead of  finding the lighting operation $L_{\mathrm{\Psi}}(\cdot)$ directly, the proposed method aims to find a transferring operation $L_{\mathrm{(\Phi\to\Psi)}}(\mathbf{X})$ that migrate the light effects (mainly the shadows) from the input to the target, which significantly reduces the difficulty of re-painting the shadows. Finally, 
a re-rendering process $P(\cdot)$ is used to combine the scene structure and light effects. The whole process can be formulated as:
\vskip -0.4cm
\vskip -0.5\baselineskip 
\begin{align}
  \mathbf{\hat{Y}} = P(L^{-1}_{\mathrm{\Phi}}(\mathbf{X}),L_{\mathrm{(\Phi\to\Psi)}}(\mathbf{X}))
\end{align}
\vskip -0.5\baselineskip 

\vskip -0.5cm
\vskip -0.5\baselineskip 
\subsection{Scene Reconversion}\label{Scene Reconversion}

\begin{figure}[t]
    \centering
    \includegraphics[width=\textwidth]{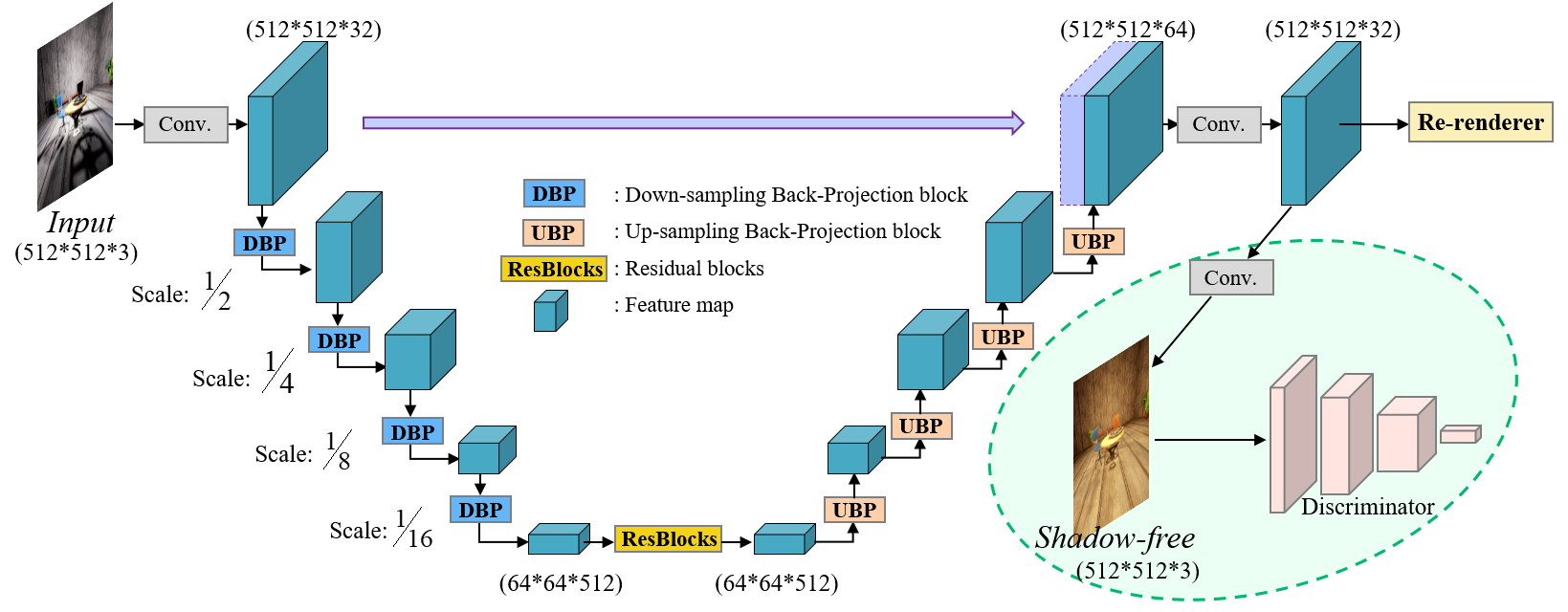}
   \vskip -1.5\baselineskip 
    \caption{Structure of the scene reconversion network. The structures in the green circle are removed after the training phase.}
    \label{fig:scene_recon}
   \vskip -0.5\baselineskip 
\end{figure}

The objective of the scene reconversion is to extract the inherent structure information from the image so that the lighting effects can be removed. As shown in Fig. \ref{fig:scene_recon}, the network adopts the auto-encoder \cite{UNet} structure with a skip connection to transfer the shallow features to the end. Firstly, the input image is down-sampled (acted by the ``DBP'' in the figure) four times to find the discriminative features (codes) for the scene. The channels are doubled after each down-sampling process to preserve information as much as possible. The features have large receptive fields, which contain much global information that benefits illumination estimation and manipulation. We design a similar auto-encoder structure as the Pix2Pix method \cite{pix2pix}, where nine residual blocks \cite{Resnet2016} (``ResBlocks'' in the figure) act to remove the light effects. Next, four blocks up-sample (acted by the ``UBP'' in the figure) the feature map back to the original size, which is then enriched by the shallow features from the skip connection. The feature map is further aggregated with a feature selection process that is acted by a convolutional layer, which reduces the channels from 64 to 32 (the top-right ``Conv."(gray rectangle) as shown in Fig. \ref{fig:scene_recon}). The feature is then sent to the following re-renderer process.

\vskip -0.5cm
\vskip -1\baselineskip 
\subsubsection{Back-Projection Block.}

\begin{figure}[t]
    \centering
    \includegraphics[width=\textwidth]{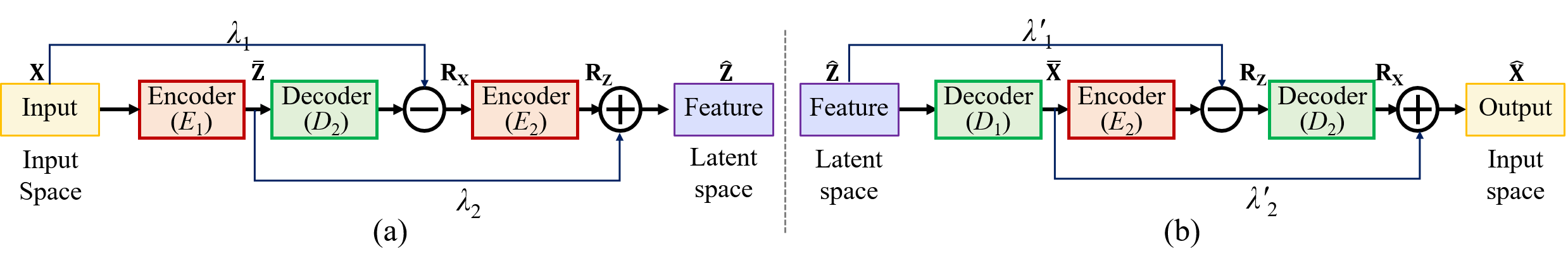}
   \vskip -1.5\baselineskip 
    \caption{Structure of the Down-sampling Back-Projection (DBP, as shown in (a)) and Up-sampling Back-Projection (UBP, as shown in (b)) blocks.}
    \label{fig:BP}
   \vskip -2\baselineskip 
\end{figure}

Instead of solely down-sampling the features with the pooling or stride-convolution process, we adopt the back-projection block that remedies the lost information through residuals. As shown in Fig. \ref{fig:BP},  the Down-sampling Back-Projection (DBP) and Up-sampling Back-Projection (UBP) blocks consist of encoding and decoding operations that map the information between the input and latent spaces. To take the DBP block for example, it firstly maps the input ($\mathbf{X}$) to latent space ($\mathbf{\overline{Z}}$) through an encoding process ($E_1$, acted by a stride convolution layer with filter size of $3\times3$, stride of 2, padding of 1). Then, a decoder ($D_2$, acted by a deconvolution layer with filter size of $4\times4$, stride of 2 and padding of 1) maps it back to the input space ($\mathbf{\hat{X}}$)to calculate the difference (residual, $\mathbf{R}_\mathbf{X}=\mathbf{X}-\mathbf{\hat{X}}$). The residual is encoded ($E_2$, acted by a stride convolution layer with filter size of $3\times3$, stride of 2 and padding of 1) to the latent space $\mathbf{R}_\mathbf{Z}$ to remedy the latent code  ($\mathbf{\hat{Z}} = \mathbf{\overline{Z}}+\mathbf{R}_\mathbf{Z}$). Mathematically, the DBP and UBP(similarly, see Fig. \ref{fig:BP}(b)) can be written as:
\vskip -0.5\baselineskip 
\begin{align}
  \hat{\mathbf{Z}} = \lambda_{2}E_{1}(\mathbf{X}) + E_{2}(D_{2}(E_{1}(\mathbf{X})) - \lambda_{1}\mathbf{X})\\
  \hat{\mathbf{X}} = \lambda_{2}D_{1}(\hat{\mathbf{Z}}) + D_{2}(E_{2}(D_{1}(\hat{\mathbf{Z}})) - \lambda_{1}\hat{\mathbf{Z}})
\end{align}

\vskip -0.5cm
\vskip -1.5\baselineskip 
\subsubsection{Semi-supervised reconversion.}

The objective of scene reconversion is to remove the light effect from the input image and construct the inherent structures. However, the ground-truth inherent structure is difficult to define because we have only the observed images. Instead of fully-supervising the network by well-defined ground-truths, it learns to estimate corresponding shadow-free images which might contain redundant information from the inherent structure. 

\begin{figure}[t]
    \centering
    \includegraphics[width=\textwidth]{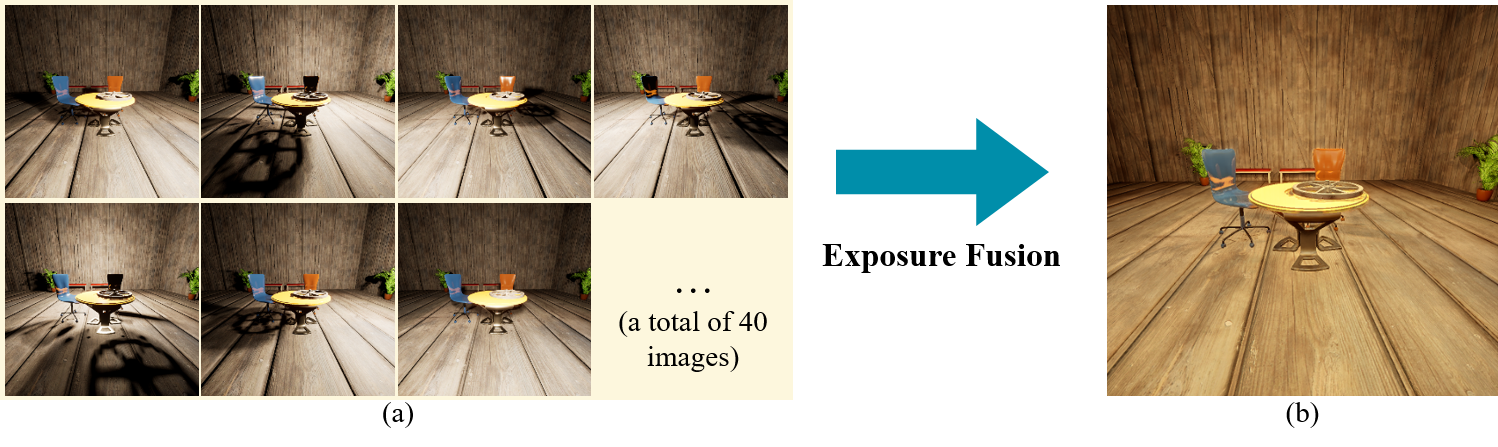}
   \vskip -1.5\baselineskip 
    \caption{An example of exposure fusion process (images with ID ``239" of the VIDIT dataset \cite{helou2020vidit}). The fourty images in (a) are captured under eight different light directions and five color temperatures of the same scene. The fusion result is shown in (b).
}
    \label{fig:HDR}
   \vskip -1\baselineskip 
\end{figure}

Exposure fusion methods \cite{HDR_opencvbase,Exposure_tom} are widely used to improve the dynamic range of the images captured in uneven light conditions. It takes several images with different exposures, and merges them to an image with better visibility. The Virtual Image Dataset for Illumination Transfer (VIDIT) dataset \cite{helou2020vidit} contains images from 390 scenes. Each scene is captured 40 times with eight different light directions and five color temperatures. Different light directions cast the shadows at different positions, which makes it possible to build shadow-free images by selecting non-shadow pixels. The same selection strategy \cite{Exposure_tom} is adopted to build shadow-free images as implemented by the OpenCV package \cite{opencv}: 1) Pixels that are too dark (underexposure) or too bright (overexposure) are given small weights. 2) Pixels with high saturation (standard deviation of RGB channels) are usually under good illumination that are given large weights. 3) Edges and textures usually contain more information and are considered more important. Fig. \ref{fig:HDR} gives an example of exposure fusion. Images in Fig. \ref{fig:HDR} (a) are captured under different light direction and color temperatures. It is obvious that these images contain shadows caused by the point-source light. After using the exposure fusion method (as shown in Fig. \ref{fig:HDR} (b)), one shadow-free image is obtained where the scene structure is obvious. The method is then used at the VIDIT dataset \cite{helou2020vidit} to generate shadow-free targets for all scenes.

\vskip -0.5cm
\vskip -0.5\baselineskip 
\subsubsection{Adversarial Learning.}
To train the scene reconversion network, a shadow-free image is formed via a convolutional layer (denoted as ``Conv." in the green circle of Fig. \ref{fig:scene_recon}), which transfers the latency structure back to the image space. However, the shadows cause holes in the input image. To fill the holes with good perceptual consistency, a discriminator is attached to assist the training of the scene reconversion network. We adopt the same discriminator structure as \cite{pix2pix} that stacks four stride-convolution layers which hierarchically extract the global representations. During the training process, the discriminator is assigned to distinguish the estimation (of the scene reconversion network) from the ground-truth shadow-free images. At the beginning, the estimation lacks structure information. The discriminator notices the weakness and makes classification based on it. At the same time, the scene reconversion network is assigned to fake the discriminator, i.e., to equip the estimation with similar structure correlation as the target shadow-free images. Mathematically, the adversarial learning is:
\begin{align}
  \mathcal{L}_{cGAN}(G,D)=\mathbb{E}_{(\mathbf{X},\mathbf{Y}_{sf})}[logD(\mathbf{X},\mathbf{Y}_{sf})]+\mathbb{E}_{\mathbf{X}}[log(1-D(\mathbf{X},G(\mathbf{X})))]
\end{align}
where the generator $G$ aims to minimize the loss $\mathcal{L}_{cGAN}(G,D)$, i.e., $G^{*}=arg\min_{G}\max_{D}\mathcal{L}_{cGAN}(G,D)$. The discriminator $D$ tries to maximize the loss $\mathcal{L}_{cGAN}(G,D)$. The term $cGAN$ indicates it is a conditional GAN structure that the discriminator has the input image $\mathbf{X}$ as prior information. Considering the estimated scene structure should be close to the ground-truth shadow-free target $\mathbf{Y}_{sf}$, the conventional L1-norm loss is used to measure the per-pixel error of the estimation. The objective for the scene reconversion network is defined as:
\begin{align}
  G^{*} = \lambda  \mathbb{E}_{(\mathbf{X},\mathbf{Y}_{sf})}[||\mathbf{Y}_{sf}-G(\mathbf{X})||]+  arg\min_{G}\max_{D}\mathcal{L}_{cGAN}(G,D)
\end{align}
where the term $\lambda$ balances the L1-norm and the adversarial losses.

\vskip -0.5\baselineskip
\subsection{Shadow Prior Estimation}
\label{shadowprior}
\begin{figure}[t]
    \centering
    \includegraphics[width=\textwidth]{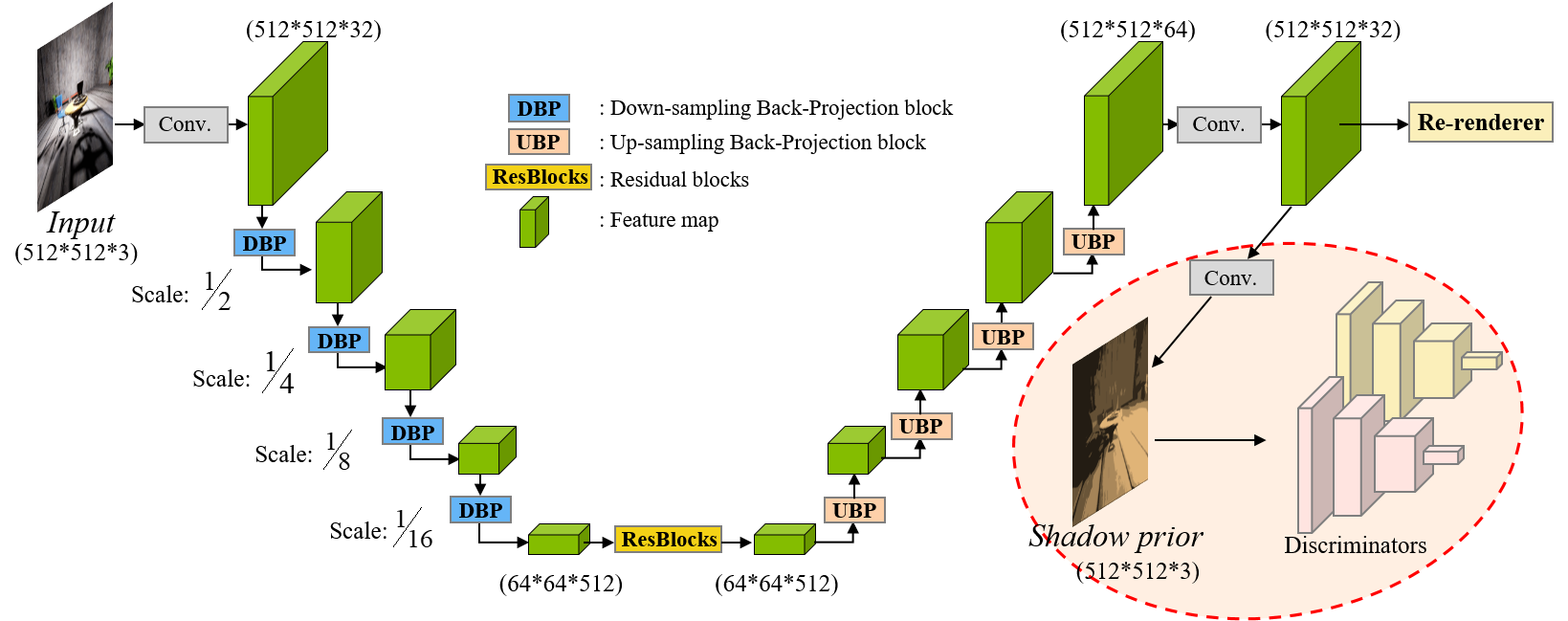}
   \vskip -1.5\baselineskip 
    \caption{Structure of the shadow prior estimation network. The structures in the red circle are removed after the training phase.
}
    \label{fig:SPN}
   \vskip -1.5\baselineskip 
\end{figure}
Different light sources cause different light effects which produce for example, different shadows and color temperatures. To produce the light effects from the target light source, we design a shadow prior estimation network with the architecture as shown in Fig. \ref{fig:SPN}. The network adopts a similar structure as the scene reconversion network (as shown in Fig, \ref{fig:scene_recon}). Specifically, there are three major modifications: 1) This shadow prior estimation network discards the skip connection, because the network gives more focus on the global light effect. The skip connection brings the local features to the output directly, which makes the network lazy to learn the global change. 2) It has another discriminator that focuses on the shadow regions. 3) The ground-truth target is the image under the target light source.  Mathematically, the objective of the shadow prior estimation network can be described as follows:
\begin{align}
  G^{*} = \lambda  \mathbb{E}_{\mathbf{X},\mathbf{Y}}[||\mathbf{Y}-G(\mathbf{X})||] +  arg\min_{G}\max_{D}\mathcal{L}_{cGAN}(G,D) \\
  + arg\min_{G}\max_{D_{shad}}\mathcal{L}_{cGAN}(G,D_{shad})
\end{align}
where $D_{shad}$ denotes the shadow-region discriminator (details will be illustrated below), and the term $\mathbf{Y}$ denotes the image under the target light source.

\vskip -0.5cm
\vskip -1\baselineskip 
\subsubsection{Shadow-region discriminator.}
\label{shad_discri}
The shadow-region discriminator adopts the same structure as \cite{pix2pix} that stacks four stride-convolution layers, which gradually extracts the global feature representations. To focus on the shadow regions, the estimation is firstly rectified to give focus to the low-intensity (dark, usually the shadows) regions through $z = min(\alpha,x)$, where the symbol $x$ denotes the estimated pixel intensity. The term $z$ represents the rectified value that will be inputted to the discriminator. The term $\alpha$ is a pre-defined threshold for the sensitivity of the shadows (empirically, it is set to $0.059=15/255$).

\vskip -0.2cm
\vskip -1\baselineskip 
\subsection{Re-rendering}
\label{rerender}
\begin{figure}[t]
    \centering
    \includegraphics[width=\textwidth]{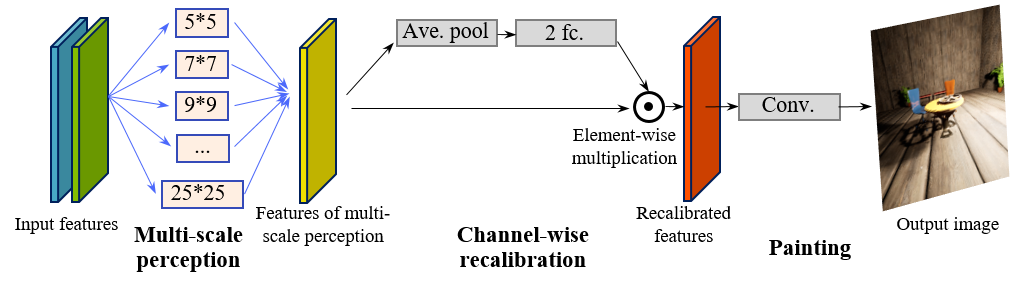}
   \vskip -1.5\baselineskip 
    \caption{Structure of the re-renderer. The inputs are two feature maps that come from the scene reconversion and shadow prior networks separately. The rectangles and cubes represent operations and feature maps respectively. 
}
    \label{fig:re-render}
   \vskip -1\baselineskip 
\end{figure}

After the processing of the scene reconversion and shadow prior estimation networks, the estimated scene structure and light effects will be fused together to produce the relighted output. As shown in Fig. \ref{fig:re-render}, the re-renderer consists of three parts: multi-scale perception, channel-wise recalibration and painting process. Both global and local information are essential for light source manipulation because global information benefits the shadow and illumination consistency, and local information enhances the details. To utilize the information of different perception scales, we propose a novel multi-scale perception block that uses filters with different perceptive sizes (e.g., filter size of $3\times3$, $5\times5$, ...), which extracts rich features for the following process. 

After processing the multi-scale perception, features with different spatial perception are merged into a single feature map, where each channel stores a type of spatial pattern. However, different patterns may have different importance for the re-rendering process. As designed in \cite{SE,DLN2020}, a recalibration process is designed to investigate the weights for different patterns, which selects the key features for the following painting process. Finally, a convolutional layer (with the filter size of $7\times7$, padding of 3, stride of 1 and a tanh activation function) paints the estimation from the feature space to the image space.
\vskip -0.5cm
\vskip -1\baselineskip 
\subsubsection{Loss function.} 
The loss function designed for the re-renderer consists of per-pixel reconstruction error and perceptual difference. The reconstruction error is measured by the wildly-used L1-norm loss. The perceptual similarity is calculated as \cite{VGG_loss} based on the features extracted from the VGG-19 network. The network is pre-trained with ImageNet dataset for image classification. The extracted features have discriminative power for visual comparison, so that they are used to measure the perceptual similarity. The loss function is defined as:
\begin{align}
 \mathcal{L}(\mathbf{Y},\hat{\mathbf{Y}})=||\mathbf{Y}-\hat{\mathbf{Y}}||+\lambda||feat(\mathbf{Y})-feat(\hat{\mathbf{Y}})||
\end{align}
where $\mathbf{Y}$ and $\hat{\mathbf{Y}}$ denote the ground-truth target and estimated images respectively. The term $feat(\cdot)$ is the feature maps extracted from the VGG-19 network. The symbol $\lambda$ is a balanced coefficient and was set to 0.01 in our experiments.

\vskip -0.5\baselineskip 
\section{Experiments}
\vskip -0.5\baselineskip 
\subsection{Implementation Details}
\vskip -0.5\baselineskip 

\subsubsection{VIDIT Dataset.}

The Virtual Image Dataset for Illumination Transfer (VIDIT) \cite{helou2020vidit} contains 390 virtual scenes with different scene contents (for example, metal, wood, etc), where there are 300 scenes for training, 45 scenes for validation and 45 scenes for testing separately. The scenes are rendered by a powerful game engine (Unreal Engine 4 \cite{UE4}) to get high-resolution images. The objective of the VIDIT dataset is for illumination manipulation. Each scene is rendered with eight light directions and five color temperatures, which results in forty images with the resolution of 1024*1024. As we mentioned in Section \ref{Scene Reconversion}, the exposure fusion method was used to generate shadow-free images for the scenes, which brings us 300 shadow-free images (work as ground-truth target) to guide the training of the scene reconversion network. We participated the ``\emph{AIM Image Relighting Challenge  - Track 1: any to one}" \cite{elhelou2020aim}. The objective is that, given an image under any types of illuminations, the method should give the estimation under a specific light source (color temperature is 4500k and light direction is from East). We used all possible pairs from the 300 training scenes to train the network, and the provided validation dataset (45 scenes) for evaluation.

\vskip -0.5cm
\vskip -1\baselineskip 
\subsubsection{Training Process.} Limited by the GPU memory and computational power, our three sub-networks (scene reconversion, shadow prior estimation and re-renderer) were trained separately. Firstly, we trained the scene reconstruction network by using the paired inputs and shadow-free targets through the designed loss functions. Similarly, we trained the shadow prior estimation network with paired input and target images. Next, we fixed the scene reconstruction and shadow prior estimation networks and removed their last convolution layer and the discriminators (the green circle in Fig. \ref{fig:HDR} and the pink circle in Fig. \ref{fig:SPN}). Finally, the re-renderer network was trained with the designed loss functions. All training images were resized from 1024*1024 to 512*512, and the mini-batch size was set to six. We used the Adam optimization method with the momentum of 0.5 and learning rate of 0.0001. The networks were randomly initialized as \cite{pix2pixHD}. As we mentioned, the scene reconstruction and shadow prior estimation networks were firstly trained independently, where each network was trained for 20 epochs. Then, the two networks were fixed, and the re-renderer network was also trained for 20 epochs. All experiments were conducted through PyTorch \cite{pytorch} on a PC with two NVIDIA GTX2080Ti GPUs. Codes have been released at {\color{blue} \url{https://github.com/WangLiwen1994/DeepRelight}}

\vskip -0.5\baselineskip 
\subsection{Analysis of the Proposed Method}
\vskip -0.5\baselineskip 
There are no evaluation methods for light source measurement, which makes it difficult to evaluate the performance of different methods. Because we have the ground-truth images under the target light condition, and we believe the estimation should be close to these ground-truth targets. Hence, the Peak Signal-to-Noise Ratio (PSNR) and Structure SIMilarity (SSIM) \cite{SSIM} are adopted to measure the similarity between the estimation and the ground-truth, where a larger value means better performance. To measure the perceptive quality, we use the Learned Perceptual Image Patch Similarity (LPIPS) \cite{LPIPS}, in which a smaller value means more perceptual similarity. 

Pix2pix \cite{pix2pix} has shown great success in the image-to-image translation tasks, like background removal, pose transfer, etc. The method is a conditional GAN structure that is trained through the adversarial strategy. The relighting problem can be regarded as an image-to-image translation task that translates the light source to the target settings. Because light source manipulation is a new topic, few methods are available for comparison.  Therefore, the Pix2Pix can be considered as the baseline model to present the efficiency of the proposed method. The Pix2Pix method is based on an auto-encoder structure where the input image is firstly down-sampled four times (scale is reduced to $1/16$), and then processed by nine residual blocks. Finally, a set of deconvolutional layers is used to up-sample the image back to the original size and the estimation is formed. Table \ref{Tab1} gives a comparison among different structures, where ShadAdv and BPAE are two variations of the Pxi2Pix network. The baseline method (Pix2Pix) achieves 16.28 dB in PSNR, 0.553 in SSIM and 0.482 in LPIPS. The performance of other structures (ShadAdv, BPAE, and the proposed DRN) will be discussed below.

\vskip -1\baselineskip 
\begin{table}
\caption{Comparison among different structures}
\vskip -1.5\baselineskip 
\label{Tab1}
\begin{center}
\begin{small}
\scalebox{1}{
\setlength{\tabcolsep}{0.8mm}{

\begin{tabular}{ccccccc}
\hline
\textbf{Method}         & \textbf{ShadAdv} & \textbf{Structure} & \textbf{Stages} & \textbf{PSNR}  & \textbf{SSIM}  & \textbf{LPIPS} \\ \hline
Pix2Pix \cite{pix2pix} & No  & Auto-Encoder   & One & 16.28 & 0.553 & 0.482          \\
ShadAdv & Yes & Auto-Encoder   & One & 17.12 & 0.569 & 0.440          \\
BPAE    & Yes & Back-Pojection & One & 17.22 & 0.573 & 0.439 \\
\textbf{DRN (proposed)} & Yes              & Back-Pojection     & Two             & \textbf{17.59} & \textbf{0.596} & \textbf{0.440}          \\ \hline
\end{tabular}
}
}
\end{small}
\end{center}
\vskip -2\baselineskip 
\end{table}

\subsubsection{Effect of the Shadow-region Discriminator.} Let us enhance the baseline, Pix2Pix, by adding the proposed shadow-region discriminator (as introduced in Section \ref{shad_discri}) to it, and entitled it as ``ShadAdv". Compared with the original Pix2Pix method, the ``ShadAdv" gives more focus on the appearance of the shadow regions. In other words, the shadow discriminator can provide better guidance for recasting the shadows of the target light source. With more accurate shadows, the PSNR is increased by 0.84($=17.12-16.28$) dB, and the perceptive quality is improved by 0.042 ($=0.482-0.440$) in terms of LPIPS.  

\vskip -0.5cm
\vskip -0.5\baselineskip
\subsubsection{Effect of the Back-Projection (BP) Block.} The ``Pix2Pix" and ``ShadAdv'' methods are based on the auto-encoder structure. As we have mentioned, it down- and up-samples the image through stacked convolutional and deconvolutional layers. The ``BPAE'' method is an enhanced version of the auto-encoder, where the down- and up-sampling processes are done by the DBP and UBP blocks (as illustrated in Fig. \ref{fig:BP}). The BP blocks are based on the back-projection theory, which remedies the lost information in the down- and up-sampling processes. Compared with the auto-encoder structure (used in ``ShadAdv''), the ``BPAE" method extracts more informative features, which enriches the structure of the estimation and increases the SSIM from 0.569 to 0.573.

\vskip -0.5cm
\vskip -0.5\baselineskip
\subsubsection{Effect of the Relighting Assumption.} 

As defined in Section \ref{assumption}, we regard the \textit{any-to-one} relighting task as a two-stage problem, where the first stage finds the scene structure $L^{-1}_{\mathrm{\Phi}}(\mathbf{X})$ and light effect $L_{\mathrm{(\Phi\to\Psi)}}(\mathbf{X})$ from the input image $\mathbf{X}$. The second stage paints $P(\cdot)$ the estimation $\mathbf{\hat{Y}}$ under the target light source. As shown in Table \ref{Tab1}, the ``Pix2Pix", ``ShadAdv'' and ``BPAE'' methods learns the mapping to the target light condition directly. The ``DRN" is the proposed method that is based on our relighting assumption. It is clear that the proposed method achieves the best reconstruction with the highest PSNR (17.59 dB) and SSIM (0.596) scores, and comparable visual similarity (0.440 of LPIPS). These suggest the effects of the proposed relighting assumption.  

\vskip -0.5\baselineskip 
\subsection{Comparison with Other Approaches}
\vskip -0.5\baselineskip 
Single image relighting is a new topic in the field of image processing. As we have mentioned, few methods are publicly available for our comparison. Besides comparing with the baseline method (Pix2Pix \cite{pix2pix}), we have also made comparisons with other representive methods.  U-Net \cite{UNet} is a popular CNN structure that was initially designed for biomedical image segmentation. It consists of down- (encoder) and up-sampling (decoder) paths to form an auto-encoder structure, where several short-connections transmit the information from the encoder to the decoder part directly. Retinex-Net \cite{RetinexNet} was designed to enlighten the low-light images based on the Retinex theory. It firstly decomposes the low-light image into the reluctance and illumination elements, and then an adjustment sub-network refines the illumination to enlighten the input images. We retrained the methods with their desired settings at the VIDIT \cite{helou2020vidit} training dataset, and the comparison was made using the VIDIT validation dataset.

\begin{table}
\vskip -1\baselineskip 
\caption{Comparison among different approaches}
\vskip -1.5\baselineskip 
\label{Tab2}
\begin{center}
\begin{small}
\scalebox{1}{

\setlength{\tabcolsep}{5mm}{
\begin{tabular}{cccc}
\hline
\textbf{Methods}        & \textbf{PSNR}  & \textbf{SSIM}  & \textbf{LPIPS} \\ \hline
U-Net \cite{UNet}                   & 16.72          & 0.616          & 0.441          \\
Retinex-Net \cite{RetinexNet}            & 12.28          & 0.162          & 0.657          \\
Pix2Pix \cite{pix2pix}                 & 16.28          & 0.553          & 0.482          \\
\textbf{DRN (proposed)} & \textbf{17.59} & \textbf{0.596} & \textbf{0.440} \\ \hline
\end{tabular}
}
}
\end{small}
\end{center}
\vskip -2\baselineskip 
\end{table}

Table \ref{Tab2} shows the results of different approaches.  Benefiting from the short-connections, the U-Net method preserves much information of the inputs, which achieves good SSIM performance. However, the short-connections preserve too much structure (detailed) information which makes the network lazy to change the light source that limits its PSNR score. The Retinex-Net method can find the inherent scene structure but fails in manipulating the light source. The proposed DRN method is able to manipulate the light source, and shows superior performance (with the best PSNR of 17.59 dB and the LPIPS score of 0.440) compared with all other methods. Also, we made use of the proposed DRN network to join the competition ``AIM2020 Image Relighting Challenge - Track 1: any to one" and have achieved the best PSNR score in the final testing phase.

\vskip -0.5cm
\vskip -1\baselineskip 
\subsubsection{Visual Comparison.} 

\begin{figure}[t]
    \centering
    \includegraphics[width=\textwidth]{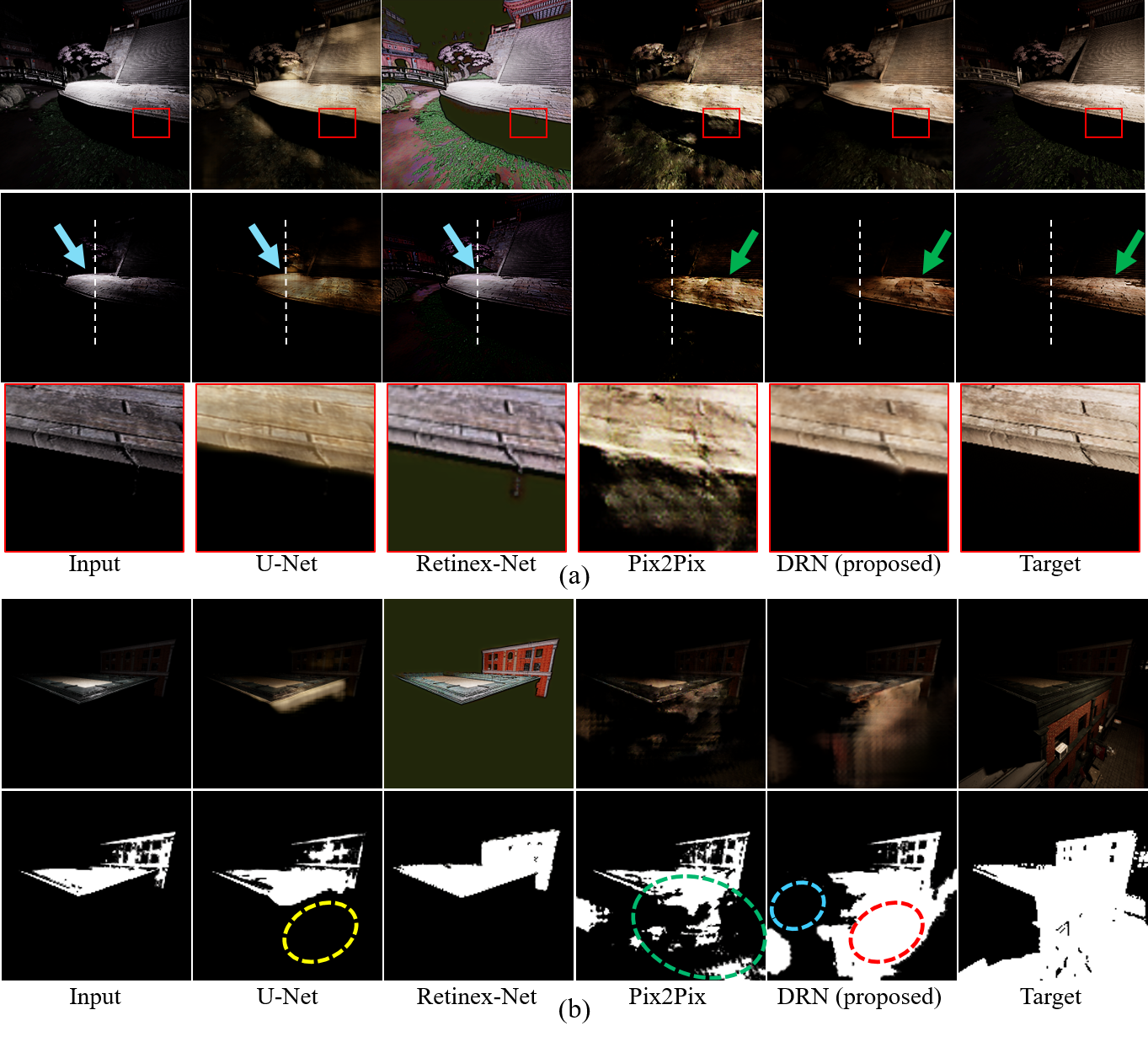}
   \vskip -1.5\baselineskip 
    \caption{Visual comparison among different approaches (zoom in for better view). Images in the $2^{nd}$ row of (a) are the results after gamma correction to highlight the light effects and the arrows indicate the light directions. The $2^{nd}$ row of (b) contains the binary mask for illuminated and shadow regions.}
    \label{fig:res}
   \vskip -1\baselineskip 
\end{figure}

Fig. \ref{fig:res} gives a visual comparison among different methods. As shown in the first row of Fig. \ref{fig:res}(a) that the Retinex-Net method \cite{RetinexNet} fails to change the color temperature where the hue of the estimation is significantly different from the others. Although U-Net \cite{UNet} produces correct color temperature for the target light source, it fails to manipulate the light direction (see the arrows in Fig. \ref{fig:res}(a)). The Pix2Pix \cite{pix2pix} can produce the correct light direction but brings in many artifacts (see the red rectangle area), which decreases the perceptual quality. The proposed method gives correct estimation for the light direction and color temperature with a good perceptual quality. 

A challenging case is shown in Fig. \ref{fig:res}(b), where the input image is nearly all black (see the top-left image of the figure). For better visualization, we provide a binary mask to show the illuminated and shadow regions (see the $2^{nd}$ row of the figure).  The U-Net \cite{UNet} fails to enlighten the building with the new light source (the yellow circle in Fig. \ref{fig:res}(b)). Pix2Pix \cite{pix2pix} brings many artifacts that cause inconsistency shadows (the green circle in Fig. \ref{fig:res}(b)). Benefited from our shadow-region discriminator, the proposed DRN enlightens the building and recasts the shadows for the new light source (the blue and red encircled regions, as shown in Fig. \ref{fig:res}(b)), which suggests superior performance as compared to all other approaches. However, the structure of the center building is completely lost (the pixels are all zero) in the input image, which makes the relighting difficult. It can be seen from the figure that the proposed method attempts to recover the color and shape of the wall, but fails to construct the detailed structures. Limited by the time and the training data, the network is designed to focus on the global light effects and lacks investigation for more advanced topics, like inpainting the building. In the future, we will continue our work and would like to invite others to work on these challenging relighting cases.

\vskip -0.5\baselineskip 
\section{Conclusion}
\vskip -0.5\baselineskip 
In this paper, we have introduced our proposed Deep Relighting Network (DRN) that achieves excellent performance in single image relighting task. We formulate the image relighting as three parts: scene reconversion, shadow prior estimation and re-rendering. We embed the back-projection theory into auto-encoder structure that significantly improves the capacity of the deep network. Benefited from adversarial learning, the proposed DRN can recast the shadows and estimate the required image from the target light source, which confirms our formulation of separating the scene and shadow structures. It is useful and can be generalized to many light-related tasks, for example, cases with dual or blind light sources, reference-based image relighting (i.e., produce images with any light source settings).  Experimental results show that the proposed DRN network outperforms all other methods. Also, it obtained the best PSNR in the competition ``AIM2020 Image Relighting Challenge - Track 1: any to one" of the 2020 ECCV conference.

\clearpage
%
%
\bibliographystyle{splncs04}
\bibliography{egbib}

\begin{thebibliography}{10}
\providecommand{\url}[1]{\texttt{#1}}
\providecommand{\urlprefix}{URL }
\providecommand{\doi}[1]{https://doi.org/#1}

\bibitem{ToDayGAN}
Anoosheh, A., Sattler, T., Timofte, R., Pollefeys, M., Gool, L.V.: Night-to-day
  image translation for retrieval-based localization. 2019 International
  Conference on Robotics and Automation (ICRA)  (2019).
  \doi{10.1109/icra.2019.8794387}

\bibitem{opencv}
Bradski, G.: {The OpenCV Library}. Dr. Dobb's Journal of Software Tools  (2000)

\bibitem{HDR97}
Debevec, P.E., Malik, J.: Recovering high dynamic range radiance maps from
  photographs. In: Proceedings of the 24th Annual Conference on Computer
  Graphics and Interactive Techniques. p. 369–378. SIGGRAPH ’97, ACM
  Press/Addison-Wesley Publishing Co., USA (1997). \doi{10.1145/258734.258884}

\bibitem{HDR_opencvbase}
Debevec, P.E., Malik, J.: Recovering high dynamic range radiance maps from
  photographs. In: ACM SIGGRAPH 2008 classes. pp. 1--10 (2008)

\bibitem{elhelou2020aim}
El~Helou, M., Zhou, R., S\"usstrunk, S., Timofte, R., et~al.: {AIM} 2020: Scene
  relighting and illumination estimation challenge. In: Proceedings of the
  European Conference on Computer Vision Workshops (ECCVW) (2020)

\bibitem{UE4}
Epic~Games, I.: Unreal {Engine} {\textbar} {The} most powerful real-time {3D}
  creation platform, \url{https://www.unrealengine.com/en-US/}

\bibitem{grammatri}
Gatys, L.A., Ecker, A.S., Bethge, M.: A neural algorithm of artistic style.
  arXiv preprint arXiv:1508.06576  (2015)

\bibitem{GAN}
Goodfellow, I., Pouget-Abadie, J., Mirza, M., Xu, B., Warde-Farley, D., Ozair,
  S., Courville, A., Bengio, Y.: Generative adversarial nets. In: Advances in
  neural information processing systems. pp. 2672--2680 (2014)

\bibitem{SR19}
{Gu}, S., {Danelljan}, M., {Timofte}, R., {Haris}, M., {Akita}, K.,
  {Shakhnarovic}, G., {Ukita}, N., {Navarrete Michelini}, P., {Chen}, W.,
  {Liu}, H., {Zhu}, D., {Xie}, T., {Yang}, X., {Zhu}, C., {Yu}, J., {Sun}, W.,
  {Tao}, X., {Deng}, Z., {Lu}, L., {Li}, W., {Guo}, T., {Shen}, X., {Xu}, X.,
  {Tai}, Y., {Jia}, J., {Yi}, P., {Wang}, Z., {Jiang}, K., {Jiang}, J., {Ma},
  J., {Liu}, Z., {Wang}, L., {Li}, C., {Siu}, W., {Chan}, Y., {Zhou}, R.,
  {Helou}, M.E., {Purohit}, K., {Kandula}, P., {Suin}, M., {A.N}, R.: Aim 2019
  challenge on image extreme super-resolution: Methods and results. In: 2019
  IEEE/CVF International Conference on Computer Vision Workshop (ICCVW). pp.
  3556--3564 (2019)

\bibitem{haris2018dbpn}
Haris, M., Shakhnarovich, G., Ukita, N.: Deep back-projection networks for
  super-resolution. In: Proceedings of the IEEE conference on computer vision
  and pattern recognition. pp. 1664--1673 (2018)

\bibitem{Resnet2016}
He, K., Zhang, X., Ren, S., Sun, J.: Deep residual learning for image
  recognition. In: Proceedings of the IEEE conference on computer vision and
  pattern recognition. pp. 770--778 (2016)

\bibitem{helou2020vidit}
Helou, M.E., Zhou, R., Barthas, J., S{\"u}sstrunk, S.: Vidit: Virtual image
  dataset for illumination transfer. arXiv preprint arXiv:2005.05460  (2020)

\bibitem{SE}
Hu, J., Shen, L., Sun, G.: Squeeze-and-excitation networks. In: Proceedings of
  the IEEE conference on computer vision and pattern recognition. pp.
  7132--7141 (2018)

\bibitem{shadow_2018}
{Hu}, X., {Zhu}, L., {Fu}, C., {Qin}, J., {Heng}, P.: Direction-aware spatial
  context features for shadow detection. In: 2018 IEEE/CVF Conference on
  Computer Vision and Pattern Recognition. pp. 7454--7462 (2018)

\bibitem{pix2pix}
Isola, P., Zhu, J.Y., Zhou, T., Efros, A.A.: Image-to-image translation with
  conditional adversarial networks. In: Proceedings of the IEEE conference on
  computer vision and pattern recognition. pp. 1125--1134 (2017)

\bibitem{EnlightenGAN}
Jiang, Y., Gong, X., Liu, D., Cheng, Y.H., Fang, C., Shen, X., Yang, J., Zhou,
  P., Wang, Z.: Enlightengan: Deep light enhancement without paired
  supervision. ArXiv  \textbf{abs/1906.06972} (2019)

\bibitem{VGG_loss}
Johnson, J., Alahi, A., Fei-Fei, L.: Perceptual losses for real-time style
  transfer and super-resolution. In: European conference on computer vision.
  pp. 694--711. Springer (2016)

\bibitem{imagenet}
Krizhevsky, A., Sutskever, I., Hinton, G.E.: Imagenet classification with deep
  convolutional neural networks. In: Advances in neural information processing
  systems. pp. 1097--1105 (2012)

\bibitem{shadow_2019}
Le, H., Samaras, D.: Shadow removal via shadow image decomposition. 2019
  IEEE/CVF International Conference on Computer Vision (ICCV) pp. 8577--8586
  (2019)

\bibitem{CNNlocalization}
{Li}, C.T., {Siu}, W.C.: Fast monocular visual place recognition for
  non-uniform vehicle speed and varying lighting environment. IEEE Transactions
  on Intelligent Transportation Systems pp. 1--18 (2020)

\bibitem{Liu2019hbpn}
Liu, Z.S., Wang, L.W., Li, C.T., Siu, W.C.: Hierarchical back projection
  network for image super-resolution. In: The Conference on Computer Vision and
  Pattern Recognition Workshop (CVPRW) (2019)

\bibitem{Liu2019abpn}
Liu, Z.S., Wang, L.W., Li, C.T., Siu, W.C.: Image super-resolution via
  attention based back projection networks. In: IEEE International Conference
  on Computer Vision Workshop (ICCVW) (2019)

\bibitem{UNet}
Long, J., Shelhamer, E., Darrell, T.: Fully convolutional networks for semantic
  segmentation. In: Proceedings of the IEEE conference on computer vision and
  pattern recognition. pp. 3431--3440 (2015)

\bibitem{SR20}
Lugmayr, A., Danelljan, M., Timofte, R.: Ntire 2020 challenge on real-world
  image super-resolution: Methods and results. In: Proceedings of the IEEE/CVF
  Conference on Computer Vision and Pattern Recognition (CVPR) Workshops (June
  2020)

\bibitem{Exposure_tom}
Mertens, T., Kautz, J., Van~Reeth, F.: Exposure fusion: A simple and practical
  alternative to high dynamic range photography. In: Computer graphics forum.
  vol.~28, pp. 161--171. Wiley Online Library (2009)

\bibitem{cGAN}
Mirza, M., Osindero, S.: Conditional generative adversarial nets. arXiv
  preprint arXiv:1411.1784  (2014)

\bibitem{Nestmeyer2020LearningPF}
Nestmeyer, T., Lalonde, J.F., Matthews, I., Lehrmann, A.M.: Learning
  physics-guided face relighting under directional light. arXiv: Computer
  Vision and Pattern Recognition  (2020)

\bibitem{seg2015}
Noh, H., Hong, S., Han, B.: Learning deconvolution network for semantic
  segmentation. In: Proceedings of the IEEE International Conference on
  Computer Vision (ICCV) (December 2015)

\bibitem{pytorch}
Paszke, A., Gross, S., Massa, F., Lerer, A., Bradbury, J., Chanan, G., Killeen,
  T., Lin, Z., Gimelshein, N., Antiga, L., et~al.: Pytorch: An imperative
  style, high-performance deep learning library. In: Advances in neural
  information processing systems. pp. 8026--8037 (2019)

\bibitem{Retinex2004}
Rahman, Z.u., Jobson, D.J., Woodell, G.A.: Retinex processing for automatic
  image enhancement. Journal of Electronic imaging  \textbf{13}(1),  100--111
  (2004)

\bibitem{VGG}
Simonyan, K., Zisserman, A.: Very deep convolutional networks for large-scale
  image recognition. arXiv preprint arXiv:1409.1556  (2014)

\bibitem{Sun2019SingleIP}
Sun, T., Barron, J.T., Tsai, Y.T., Xu, Z., Yu, X., Fyffe, G., Rhemann, C.,
  Busch, J., Debevec, P.E., Ramamoorthi, R.: Single image portrait relighting.
  ACM Transactions on Graphics (TOG)  \textbf{38},  1 -- 12 (2019).
  \doi{10.1145/3306346.3323008}

\bibitem{DLN2020}
{Wang}, L.W., {Liu}, Z.S., {Siu}, W.C., {Lun}, D.P.: Lightening network for
  low-light image enhancement. IEEE Transactions on Image Processing
  \textbf{29},  7984--7996 (2020). \doi{10.1109/TIP.2020.3008396}

\bibitem{pix2pixHD}
Wang, T.C., Liu, M.Y., Zhu, J.Y., Tao, A., Kautz, J., Catanzaro, B.:
  High-resolution image synthesis and semantic manipulation with conditional
  gans. In: Proceedings of the IEEE Conference on Computer Vision and Pattern
  Recognition (2018)

\bibitem{SSIM}
Wang, Z., Bovik, A.C., Sheikh, H.R., Simoncelli, E.P.: Image quality
  assessment: from error visibility to structural similarity. IEEE transactions
  on image processing  \textbf{13}(4),  600--612 (2004)

\bibitem{RetinexNet}
Wei, C., Wang, W., Yang, W., Liu, J.: Deep retinex decomposition for low-light
  enhancement. ArXiv  \textbf{abs/1808.04560} (2018)

\bibitem{HDR_ECCV}
Wu, S., Xu, J., Tai, Y.W., Tang, C.K.: Deep high dynamic range imaging with
  large foreground motions. In: Proceedings of the European Conference on
  Computer Vision (ECCV) (September 2018)

\bibitem{HE_1}
{Yadav}, G., {Maheshwari}, S., {Agarwal}, A.: Contrast limited adaptive
  histogram equalization based enhancement for real time video system. In: 2014
  International Conference on Advances in Computing, Communications and
  Informatics (ICACCI). pp. 2392--2397 (2014)

\bibitem{segYu}
Yu, C., Gao, C., Wang, J., Yu, G., Shen, C., Sang, N.: Bisenet v2: Bilateral
  network with guided aggregation for real-time semantic segmentation. arXiv
  preprint arXiv:2004.02147  (2020)

\bibitem{LPIPS}
Zhang, R., Isola, P., Efros, A.A., Shechtman, E., Wang, O.: The unreasonable
  effectiveness of deep features as a perceptual metric. In: Proceedings of the
  IEEE conference on computer vision and pattern recognition (2018)

\bibitem{portrait_relight_2019}
Zhou, H., Hadap, S., Sunkavalli, K., Jacobs, D.W.: Deep single-image portrait
  relighting. In: Proceedings of the IEEE/CVF International Conference on
  Computer Vision (ICCV) (2019)

\end{thebibliography}
\end{document}